\documentclass{article}
    \PassOptionsToPackage{numbers, compress}{natbib}

    \usepackage[preprint]{neurips_2020}

\usepackage[utf8]{inputenc} 
\usepackage[T1]{fontenc}    
\usepackage{hyperref}       
\usepackage{url}            
\usepackage{booktabs}       
\usepackage{amsfonts}       
\usepackage{nicefrac}       
\usepackage{microtype}      

\usepackage{amsmath}
\usepackage{times}
\usepackage{latexsym}
\usepackage{graphicx}
\usepackage{float}
\usepackage{caption}
\usepackage{subcaption}
\usepackage{stackengine}
\usepackage{xcolor}
\usepackage[normalem]{ulem}
\useunder{\uline}{\ul}{}
\usepackage{multirow}
\usepackage{makecell}

\newcommand\reals{\rm I\!R}
\title{SEAL: Segment-wise Extractive-Abstractive Long-form Text Summarization
}

\author{Yao Zhao \\
  Google Research, Brain Team \\
  \texttt{yaozhaoyz@google.com} \\\And
  Mohammad Saleh \\
  Google Research, Brain Team \\
  \texttt{msaleh@google.com} \\\And
  Peter J. Liu \\
  Google Research, Brain Team \\
  \texttt{peterjliu@google.com} \\
  }

\date{}

\begin{document}
\maketitle
\begin{abstract}
Most prior work in the sequence-to-sequence paradigm focused on datasets with input sequence lengths in the hundreds of tokens due to the computational constraints of common RNN and Transformer architectures. In this paper, we study long-form abstractive text summarization, a sequence-to-sequence setting  with input sequence lengths up to 100,000 tokens and output sequence lengths up to 768 tokens. We propose \emph{SEAL}, a Transformer-based model, featuring a new encoder-decoder attention that dynamically extracts/selects input snippets to sparsely attend to for each output segment. Using only the original documents and summaries, we derive proxy labels that provide weak supervision for extractive layers simultaneously with regular supervision from abstractive summaries. The SEAL model achieves state-of-the-art results on existing long-form summarization tasks, and outperforms strong baseline models on a new dataset/task we introduce, \emph{Search2Wiki}, with much longer input text. Since content selection is explicit in the SEAL model, a desirable side effect is that the selection can be inspected for enhanced interpretability.
\end{abstract}

\section{Introduction}

Text summarization is a language generation task that seeks to output concise and informative content given possibly multiple input documents.
Abstractive summarization aims to summarize text beyond solely copying text segments (usually whole sentences), using novel words and phrases to be more concise, comprehensive, or achieve a certain style.
Neural abstractive summarization \citep{Rush2015ANA, Nallapati_2016} is a data-driven summarization approach that trains sequence-to-sequence \citep{sutskever2014sequence} models on large numbers of document-summary pairs,  and have demonstrated promising results \citep{pointer_generator, bertsum, bottomup, unilm} on summarizing relatively short texts where the input documents typically contain up to 500-1000 document tokens and up to 100-200 summary tokens (a few sentences or a paragraph).

Despite progress in relatively short-form settings, the development of long-form neural abstractive summarization (LF-NAS) was constrained due to the lack of large-scale long-form datasets.
Recently, there has been increasing interest in collecting such datasets for single-document (SDS),
including for scientific articles from arXiv/PubMed \citep{arxiv}, and multi-document summarization (MDS) \citep{wikisum},
although the challenge remains of developing model architectures that can cope with that increased
data scale, in particular sequence length.

Prior work on LF-NAS typically divided the task into two separate stages, mainly the result of memory and computation constraints associated with processing
long input sequences with RNN \citep{Hochreiter:1997:LSM:1246443.1246450,chung2014empirical} or Transformer models \citep{transformer}.
The first, extractive stage selects a small fraction of the input text
for the second abstractive stage, which
typically relies on some flavor of sequence-to-sequence model
to process extracted inputs into summaries. 

In this work, we systematically study single models for LF-NAS trained jointly with weak supervision for extractive layers and with regular supervision for abstractive layers.  
To use a single approach for both SDS and MDS we break input documents into sequences of snippets 
(Section~\ref{sec:snippet})
and consider three general categories of modeling approach: the first truncates inputs to leading snippets (Section \ref{sec:trunc}); the second, compressive-abstractive, compresses all snippets into a 
fixed-length representation that is used as input to the abstractive attention layers (Section \ref{sec:ca});
the third, extractive-abstractive, encodes each snippet independently,
then sparsely selects snippets when applying abstractive attention layers (Section \ref{sec:ea}). 
The sparse selection is performed by a scorer sub-network and gating function. 
The scorer is weakly supervised by proxy extractive labels automatically calculated from input snippets and gold summaries similar to \citet{Nallapati_summarunner}.
The abstractive attention layers employ full encoder-decoder attention on the resulting input representations
along with self-attention, which is supervised using the abstractive summaries.
During training, the extractive and abstractive losses are 
added together and optimized. 

We propose the Segment-wise Extractive-Abstractive Long-form model (Section \ref{sec:seal}), or \emph{SEAL}, that generalizes and improves the extractive-abstractive approach by augmenting the 
scorer inputs with previously decoded tokens for each segment, allowing the model to dynamically select input
snippets during decoding. Although interpretability was not our goal, as a nice side-effect the model shows the link between selected 
input snippets and decoded segments, resulting in better interpretability than 
soft attention models.

We are able to apply our models to very long input documents and achieve state-of-the-art results with the SEAL model
on the arXiv/PubMed datasets. To further demonstrate the advantage of the SEAL model over other approaches,
we collected a massive, new dataset, \emph{Search2Wiki}, which we use to generate full Wikipedia pages from top search results, with up to 100,000 input tokens.

\section{Related Work}

Sequence-to-sequence has become a dominant paradigm in abstractive summarization using encoder-decoder architectures
based on RNNs and more recently Transformer \citep{transformer}. Transformers have shown to be more effective at
incorporating long-term dependencies \citep{wikisum,gpt2} than RNNs but have issues
scaling, running into computation and memory limitations
quickly beyond 1000 sequence tokens ($O(n^2)$ in sequence length $n$). One option is to simply truncate sequences, but 
depending on the task, may drop important information. 

Another option is to factor the summarization problem into extractive and abstractive stages.
The Extract-Abstract framework has received increased attention recently \citep{wikisum,li-etal-2018-guiding,gehrmann-etal-2018-bottom, tlm,heirachical_transformer} as a natural way to reduce the input size for abstractive sequence-to-sequence models. In past work
the abstractive model was trained separately from the extractive stage.

\citet{chen-bansal-2018-fast} trained extractor and abstractor sub-modules separately then used REINFORCE \citep{reinforce}
to fine-tune the non-differentiable extractive selection end-to-end. However,
the REINFORCE slows training down significantly and is not scalable for our setting.

\citet{amplayo2019informative} proposed encoding documents using an encoder derived from a pre-trained
autoencoder followed by a pooling layer before processing with a separately trained decoder, 
which they call Condense-Abstract.

Our proposed SEAL model and baseline models loosely fit in the Extract-Abstract or Condense-Abstract frameworks, except
they are trained jointly in a single model end-to-end and without pre-training. 

For unconditional language modeling,
\citet{child2019sparsetransformer} investigated scaling Transformer by using sparse attention, reducing
complexity to $O(n \sqrt[p]{n})$; \citet{kitaev2020reformer} replaced dot-product attention by one that uses locality-sensitive hashing, changing its complexity to $O(nlog(n))$; \citet{roy2020efficient} endowed self-attention with a sparse routing module based on online k-means while reducing its complexity to $O(n^{1.5})$; while \citet{Dai_2019,grave2016improving} cache state across
attention frames to incorporate longer context.

Pointer-networks \citep{pointer_networks} are supervised to copy input tokens to the output, whereas
in our extractor/scorer it is supervised to provide input to the subsequent abstractor layers.

\citet{shazeer2017outrageously,Gross_2017} used a gating network to save computation in very large
networks; our scorer/gate mechanism also saves computation and memory by restricting attention to small subsets of the input data.

In our model, the extraction can be interpreted similarly to hard attention, which restricts 
what the downstream model layers (decoder) see. Hard attention has been designed
into sequence models previously
to increase the interpretability of predictions \citet{lei-etal-2016-rationalizing}.

Pre-training sequence-to-sequence models using massive external data using a self-supervised objective as in \citet{2019t5, song2019mass, unilm, zhang2019pegasus}
has lead to improvements in downstream summarization tasks, although this line of work is orthogonal to our focus on scaling the Transformer to long inputs.
Thus we train from randomly initialized weights.

\citet{sauper-barzilay-2009-automatically} generated Wikipedia extractively 
from search results for a few categories. \citet{wikisum} augmented search results
with cited references and generated articles (focusing on lead section) abstractively
with sequence-to-sequence architectures. Our work differs from the latter
by focusing on generating full articles and using only search results, although many more (75 vs 10).

\section{Models}

\begin{figure}[tbh]
\centering
\includegraphics[width=0.8\textwidth]{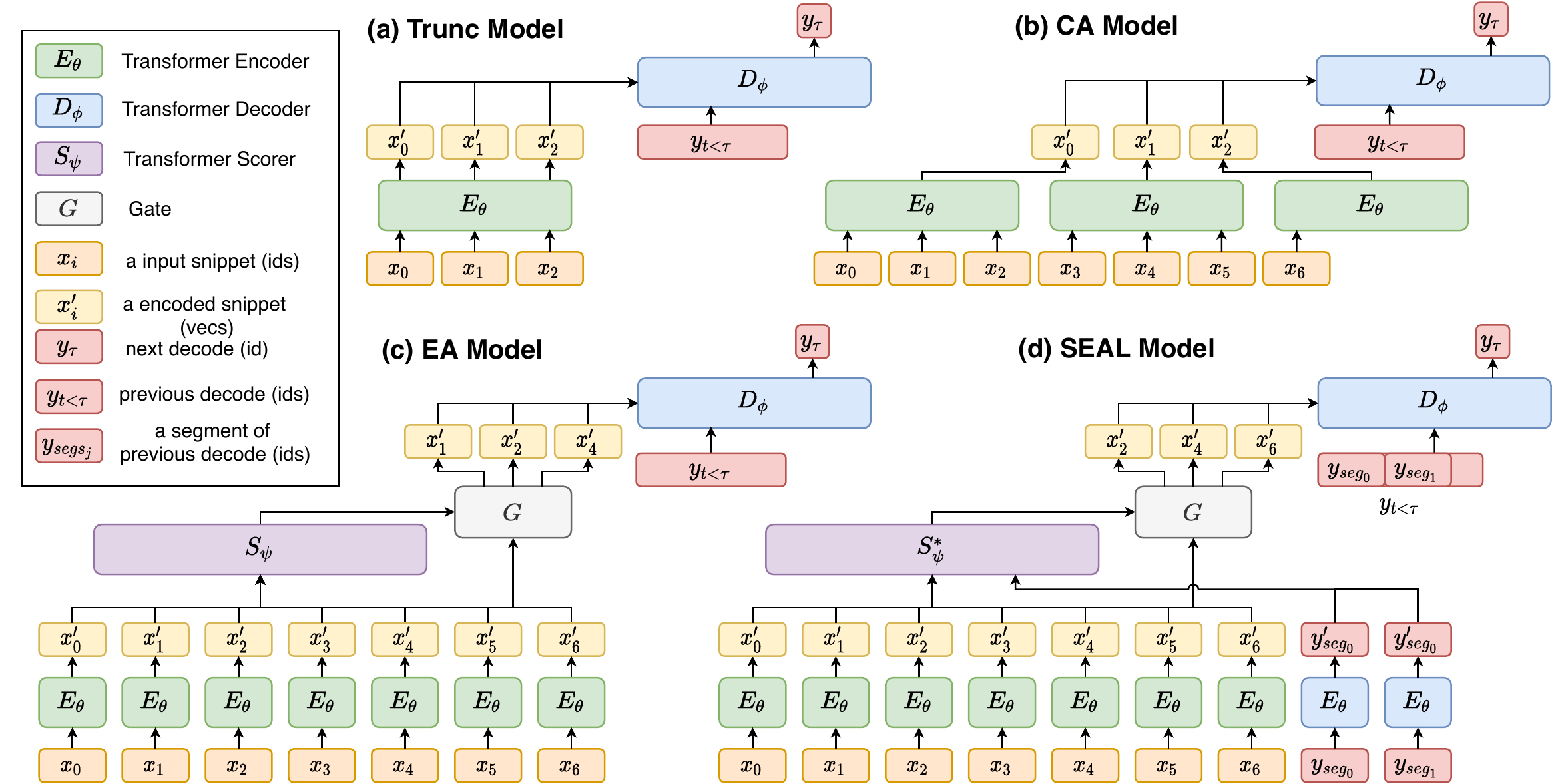}
\caption{Model architectures. $E_\theta$, $D_\phi$, $S_\psi$, $S_\psi^*$ are encoder, decoder and scorers that contains trainable parameters $\theta$, $\phi$, $\psi$. $G$ is the gating function that selects and concatenates top scored snippet representations up to a certain maximum length. $x_{i}$ are inputs snippet IDs (each $x_i$ is a sequence of IDs),
$x'_{i}$ are encoded/compressed representations of input snippets. $y_\tau$, $y_{t<\tau}$, $y_{seg_j}$ are current decode IDs, all previous decode IDs, and previous decode IDs in segment j.
In this figure, there are in total 7 inputs snippets and decoders always attend up to 3 input representations, the SEAL model is decoding the third segment.
}
\label{fig:arch}
\end{figure}

\begin{figure}[tbh]
\centering
\includegraphics[trim=0 15 0 0, clip, width=0.3\textwidth]{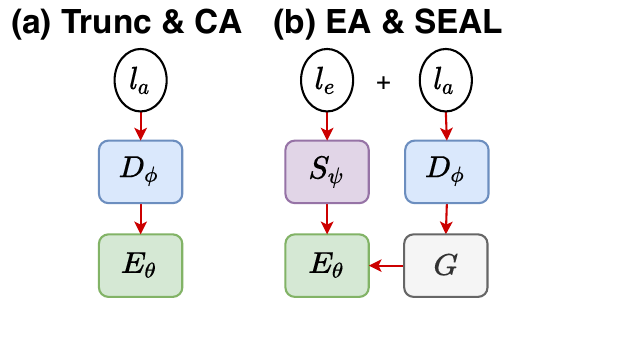}
\caption{Losses and how gradients flow. The Left side are Trunc and CA model. The right side are EA model and SEAL model. $l_a$ and $l_e$ are abstractive and extractive loss, red arrows are gradients.
}
\label{fig:loss}
\end{figure}

\begin{figure*}[tbh]
\centering
\includegraphics[trim=100 70 150 20,
clip,width=0.7\textwidth]{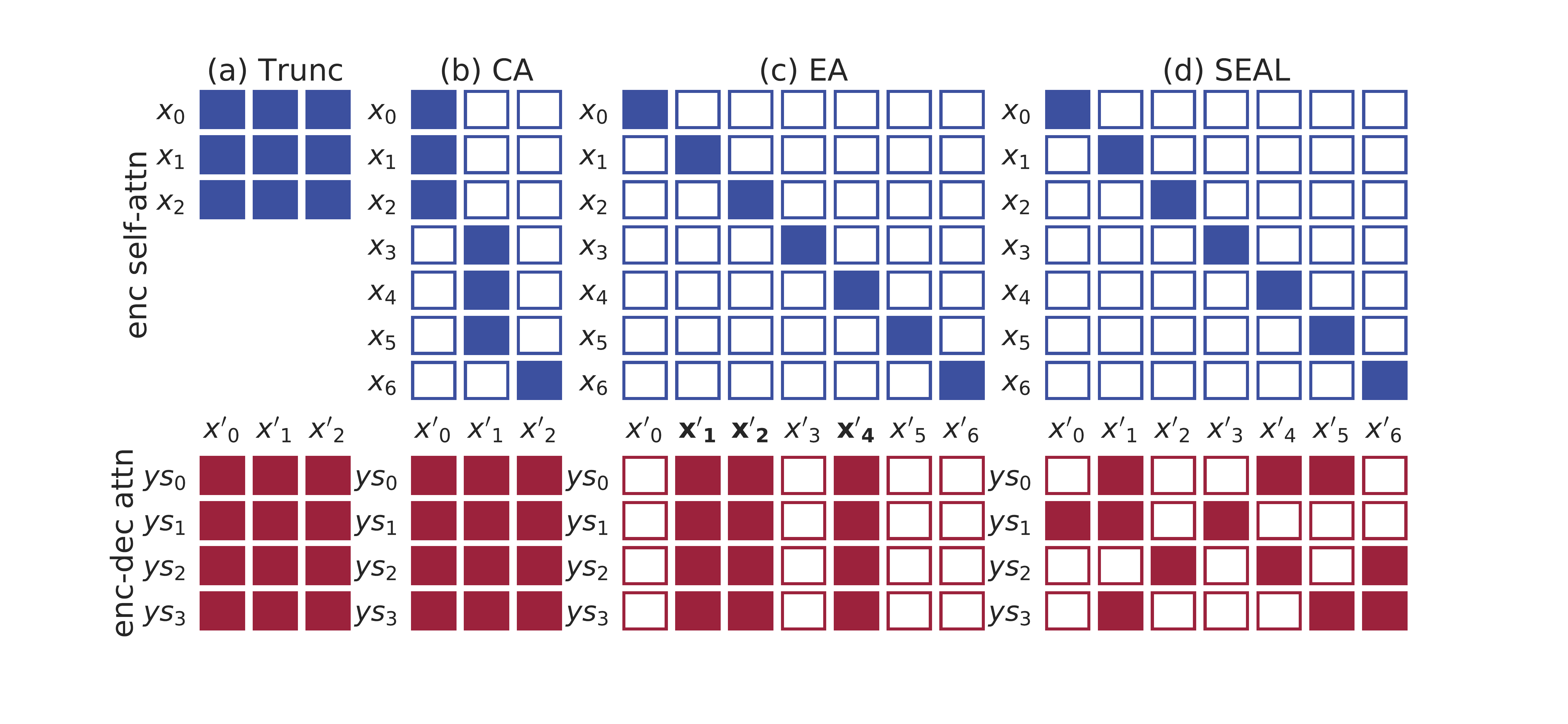}
\caption{Illustration of encoder self-attention and encoder-decoder attention maps for four models considered. $x_{i}$ are inputs snippets (encoders' inputs),
$x'_{i}$ are encoded/compressed representations (encoders' outputs, decoders' inputs) that correspond to input snippets. $ys_j$ are decode segments (decoder's outputs) representing parts of the long decode sequence. Encoder self-attentions from $x_{i}$ to $x'_{i}$ are colored in blue and encoder-decoder attentions from $x'_{i}$ to $ys_{j}$ are colored in red. Note each square represents a sequence of tokens in a input snippet or a decode segment, not a single token. 
}
\label{fig:attention}
\end{figure*}

In this section, we discuss architectures (Fig.~\ref{fig:arch}), losses (Fig.~\ref{fig:loss}) and attention maps (Fig.~\ref{fig:attention}) of  three general approaches and our proposed method to deal with LF-NAS: \textbf{truncated input} (Section \ref{sec:trunc}), \textbf{compressive-abstractive} (Section \ref{sec:ca}), \textbf{extractive-abstractive} (Section \ref{sec:ea}) and \textbf{SEAL} (Section \ref{sec:seal}), our proposed more general form of extractive-abstractive model.
We encode the text using sub-word tokenization similar to \citep{wu2016googles} with a vocabulary size of $32,776$.
Each model generates an output sequence IDs $y_\tau$ from a sequence of input snippets IDs $\{x_i\}$ (Section~\ref{sec:snippet}).
A snippet is defined as a continuous text-span, such as one or few sentences or a paragraph.
All our models' components are based on Transformer \cite{transformer}. 
A Transformer encoder maps an input sequence of tokens to a sequence of high-dimensional vectors in $\reals^d$ using self-attention and feed-forward layers.
A Transformer decoder auto-regressively (by self-attention) generates an output sequence of tokens 
while attending to this input representation (by encoder-decoder cross attention).

\subsection{Unified Approach for Single and Multi-Document Summarization}
\label{sec:snippet}
To unify our approach to SDS and MDS, we break a input document or a list of documents into snippets with the following criteria:
(1) concatenate tokens in continuous sentences until reaching a \emph{maximum snippet length}, $L_{snpt}$. This helps to reduce the percentage of paddings (empty tokens) in each snippet for better compute efficiency;
(2) in the unlikely case of a single sentence exceeding the maximum length, truncate that sentence;
(3) ensure that snippets do not span across document boundaries;
(4) order by their natural order of appearances within each document. In MDS, they are ordered by the order in the data (Section ~\ref{sec:data/wiki});
(5) stop adding snippets when their number reaches \emph{maximum snippets}, $N_{snpt}$.

\subsection{Truncated Input Model (Trunc)}
\label{sec:trunc}

One limitation of a standard Transformer model is it does not scale well to longer inputs
due to the $O(n^2)$ complexity of encoder self-attention to input length.
We truncate the input sequences as in \citet{wikisum} to a \emph{maximum input length}, $L_{input}$, only including the leading few snippets, Fig.~\ref{fig:arch}(a). 
We refer this model as \emph{Trunc}.
\begin{equation*}
p(y_\tau) = D_{\phi}(y_{t < \tau}, E_{\theta}(\{x_i\}))
\label{eq:trunc}
\end{equation*}
Encoder $E_\theta$ and decoder $D_\phi$, parameterized by $\theta$ and $\phi$ respectively.
As shown in Fig.~\ref{fig:loss}(a), the Trunc model is trained using the standard teacher-forcing and cross-entropy loss over all generated tokens $l_a =\nolinebreak \sum_{y\tau}^{T}\hat{y_\tau} log(p(y_\tau))/T$. $\hat{y_\tau}$ is the target decode sequence and $T$ is the length of decode.
The few snippets have full self-attention among themselves and the decoder has full attention over the snippets, Fig.~\ref{fig:attention}(a).

\subsection{Compressive-Abstractive (CA) Model}
\label{sec:ca}

The second approach, \emph{compressive-abstractive}, encodes and compresses continuous snippets into shorter representations,
and concatenates all representations as the decoder input, Fig.~\ref{fig:arch}(b).
\begin{equation*}
p(y_\tau) = D_{\phi}(y_{t < \tau}, \{C_{\theta}(x_i)\})
\label{eq:ca}
\end{equation*}
where $C_\theta$ is the Transformer encoder that also compresses the input snippets. The CA model is trained with a similar loss $l_a$ as the Trunc model, Fig.~\ref{fig:loss}(a).

\citet{amplayo2019informative} pooled each input representation $\mathbf{x}^{l \times d}$ into a single vector $\mathbf{x}^{d}$ ($l$ is the sequence length and $d$ is the representation dimension) whereas we
compress a snippet group into a short sequence of vectors $\mathbf{x}^{c \times d}$ for richer representation ($c$ is the compressed size).
A \emph{snippet group} is a block of continuous $k$ snippets for SDS or all snippets within the same document for MDS.
The compression is implemented by concatenating learnable vectors to Transformer encoders' inputs \cite{compression} and retrieving the  processed vectors as compressed representations.
As shown in Fig.~\ref{fig:attention}(b), the compressed representation is derived from full self attention to snippets within the compression group, and the decoder has full attention to all compressed representations.

\subsection{Extractive-Abstractive (EA) Model}
\label{sec:ea}
The third approach, \emph{extractive-abstractive}, first encodes each input snippet separately with $E_\theta$, 
assigns scores to encoded snippets with scorer $S_\psi$ (a modified Transformer encoder),
then selects encoded snippets by scores through gating function $G$. The decoder attends to sequences selected by $G$, Fig.~\ref{fig:arch}(c).
\begin{equation*}
p(y_\tau) = D_{\phi}(y_{t < \tau}, G(\{E_{\theta}(x_i)\},S_{\psi}(\{E_{\theta}(x_i)\}))
\label{eq:ea}
\end{equation*}

Each input snippet only has encoder self-attention to itself, 
the decoder has attention to selected snippets through the gating function, Fig.~\ref{fig:attention}(c).


The \textbf{scorer $S_\psi$} utilizes a Transformer encoder to map a list of input snippets representations $\{\mathbf{x}_i^{l \times d} \}_n$ to a list of scores $\{s_i\}_n$ ($l$ is the sequence length, $d$ is the representation dimension and $n$ is the number of snippets). It consists of a attention-pooling layer, Transformer encoder, and a feed-forward layer.  The attention-pooling layer \citet{wang-etal-2017-gated} reduces snippets representations from $\{\mathbf{x}_i^{l\times d}\}_n$ to $\{\mathbf{x}_i^{d}\}_n$. The Transformer encoder process the concatenation of pooled snippets $\mathbf{x}^{n\times d}$ for contextual representation across snippets. The feed-forward layer assigns a scores $s_i$ for each contextual snippets representation $\mathbf{x}_i^{d}$.
In MDS, we assign a document id to each snippet, and add learnable document embedding to the pooled snippet representation. 


After each snippet is assigned a score by the scorer, we apply a \textbf{gating function}, $G$, to select snippets based on their scores. It is implemented in the following way: (1) sort the snippets by their predicted scores; (2) concatenate each snippet representation until their total length reaches a limit. We refer to this length limit as \emph{maximum extractive length}, $L_{ext}$. Note concatenation is done by matrix multiplication of the sorting mask (an one-hot matrix mapping inputs positions to sorted concatenated positions) and encoded snippets, thus gradients can back-propagate through the gating function to the encoder.


The encoder $E_\theta$, scorer $S_\psi$ and decoder $D_\phi$ are \textbf{jointly trained} with two losses, $l_a$ and $l_e$. The \textbf{abstractive loss $l_a$} is the same as Trunc and CA model (Section \ref{sec:trunc}).
The \textbf{extractive loss $l_e$} provides supervision for the scorer and encoder to assign higher scores to better snippets for the decoder to attend to.
During training, we calculate text similarities between gold summary and input snippets as weakly supervised proxy labels $\{l_i\}_n$, and minimize the $l2$ distance between model predicted scores and proxy labels, $l_e = \sum_i^n (s_i - l_i)^2/n$.

We minimize the sum of the two losses $l_a$ and $l_e$ during training as shown in Fig.~\ref{fig:loss}(b). The $l_a$ loss back-propagates to $D_\phi$ and $E_\theta$ through $G$, while the $l_e$ loss back-propagates to $S_\psi$ and $E_\theta$. This differs from two-stage models \citep{tlm, heirachical_transformer} where the extractive and abstractive stages use different encoders and trained separately.

\subsection{Segment-wise Extractive-Abstractive Long-form (SEAL) Model}
\label{sec:seal}
The SEAL model encodes snippets in the same way as the EA model. On the decoder side, the model divides the process into non-overlapping segments, each with size \emph{segment length}, $L_{seg}$, Fig.~\ref{fig:arch}(d). Different snippets are selected for the decoder to attend to at each decode segment. 
\begin{equation*}
p(y_\tau) = D_{\phi}(y_{t < \tau}, G(\{E_{\theta}(x_i)\},S_{\psi}^*(\{E_{\theta}(x_i)\}, y_{t < seg(\tau)}))
\label{eq:sea}
\end{equation*}
where $seg(\tau) = floordiv(\tau, s) \times s$ is current decoding step $\tau$'s 
segment (with size $s$) starting index.
The inputs to the scorer $S_\psi^*$ are all encoded snippets $\{E(x_i)\}$ (unchanged) and prior decode segments $y_{t < seg(\tau)}$ (changed at the start of each decode segment).
The gating function $G$ selects subsets of snippets based on $S_\psi^*$ assigned scores at each segment.

This model has the same self-attention mask as the EA model and an encoder-decoder attention that dynamically changes between decode segments as shown in Fig.~\ref{fig:attention}(d). The dynamic selection of snippets allows more efficient usage of attention memory, more targeted  proxy labels (Section~\ref{sec:exp}), and improved interpretability (Section~\ref{sec:interp}).

At the start of each decode segment, the encoder $E_\theta$ encodes the tokens of each $k$ previous segment $\{\mathbf{y}_{seg_{j}}^s\}_k$ into representations $\{\mathbf{y}_{seg_{j}}^{s\times d}\}_k$ (k is the number of previous decode segments). 
The \textbf{segment-wise scorer} consists of an attention-pooling layer and a transformer encoder. 
The attention-pooling layer pools $\{\mathbf{y}_{seg_{j}}^{s\times d}\}_k$ and concatenates them to $\mathbf{y}^{d\times k}$. 
The transformer encoder $S_\psi^*$ not only processes the pooled input snippets $\mathbf{x}^{n\times d}$ (by self-attention) same as in the EA model, but also attends to  $\mathbf{y}^{d\times k}$ (by encoder-decoder cross attention) such that the scorer is aware of what has been decoded.  (Refer to “attention\_layer” in BERT Tensorflow library\footnote{https://github.com/google-research/bert/blob/master/modeling.py}, the “from\_tensor” is pooled snippet representations, the “to\_tensor” is pooled previous decode segments.)


Training the SEAL model is similar to the EA model, Fig.~\ref{fig:loss}(b), except: (1) The number of supervised labels increases from $n$ to $n \times m$, where $n$ is the number of input snippets and $m$ is the number of decode segments. The proxy labels are calculated as similarities between each gold summary segment and input snippet. Thus, the extractive loss is $l_e = \sum_i^n\sum_j^m (s_{ij} - l_{ij})^2 / mn$. (2) Decoding segments are trained in parallel while attending to different inputs snippets.

The SEAL model is a more general version of the EA and Trunc models.
When $L_{seg}=L_{dec}$,  the SEAL model reduces to an EA model. When $L_{ext}=L_{input}$, the EA model reduces to a Trunc model.

\section{Datasets}
\label{sec:data/wiki}

\begin{table}[tbh]
\centering
\caption{Statistics of the long-text summarization tasks. The lengths are calculated in the number of subword tokens, 1 word on average equals to 1.3 subword. }
\label{tab:data}
\resizebox{0.5\columnwidth}{!}{%
\begin{tabular}{c|c|cc|cc}
\hline

\multirow{2}{*}{dataset} & \multirow{2}{*}{examples} & \multicolumn{2}{c|}{input length} & \multicolumn{2}{c}{target length}  \\ 
 &  & mean  & 95\% & mean & 95\% \\ \hline
Search2Wiki & 4.8M & 37k & 90k &  310 & 510  \\ 
Arxiv& 215k  & 10k  & 24.2k & 460  & 754  \\ 
Pubmed   & 133k  & 4.4k & 10.2k & 331 & 538  \\ 

\hline
\end{tabular}
}

\end{table}

Several existing datasets are suitable as benchmarks for LF-NAS (statistics in Table.~\ref{tab:data}).
\citet{arxiv} collected scientific publications from arXiv and PubMed and used articles' bodies to generate abstracts.
To visualize selections by the SEAL model, 
we use a popular (relatively) short summarization dataset CNN/DailyMail\cite{cnndailymail}, non-anonymized version as in \citet{pointer_generator}, for ease of presentation. It contains newspaper articles paired with bullet point summaries.

\citet{wikisum} approached generating the lead section of English Wikipedia article  as  multi-document summarization from reference documents (Wikipedia references and web search results) and introduced the WikiSum dataset. 
To demonstrate the performance of our models on much longer input and target sequences we created a similar LF-NAS dataset named \emph{Search2Wiki}. The dataset consists of full English Wikipedia articles as the target summaries and a collection of search result documents about the topic as input documents. The main differences are: 1) the input uses the top 75 search result documents compared to the top 10 in WikuSum. This is important to demonstrate the effectiveness of our proposed models on longer input sequences; 2) We drop Wikipedia references which could vary a  lot per page. It allows generating pages for entities not currently in Wikipedia; 3) To make the dataset more abstractive, we apply a stronger filtering of Wiki-clone documents retrieved from search, please refer to Appendix ~\ref{web2wiki_clone} for more details.
The order of documents presented to model is by their search relevance in descending order.

\section{Experiments and Results}
\label{sec:exp}

\paragraph{Experimental details}
Our Transformer block \citep{transformer} has hidden size $d_{model}=512$, feed forward size $d_{ff}=512$ and attention heads $h=8$.
We set number of Transformer layers to 2 in encoder, 1 in scorer, and 6 in decoder, making the total number of parameters $223M$ for the SEAL model (Trunc, CA and EA model have similar number of parameters).
All models are trained with adafactor ~\citep{shazeer2018adafactor} optimizer, batch size of 256, dropout rate of 0.1, 
learning rate of 0.01 with square root decay. We train the models until perplexity stopped decreasing on dev set (200k steps for arXiv, PubMed and 400k steps for Search2Wiki).
In order to have a clear comparison between models, all our models are decoded with greedy decoding.

\begin{figure}[tbh]
\centering
\begin{subfigure}[t]{0.39\textwidth}
\includegraphics[trim=50 0 50 50,clip,width=\linewidth]{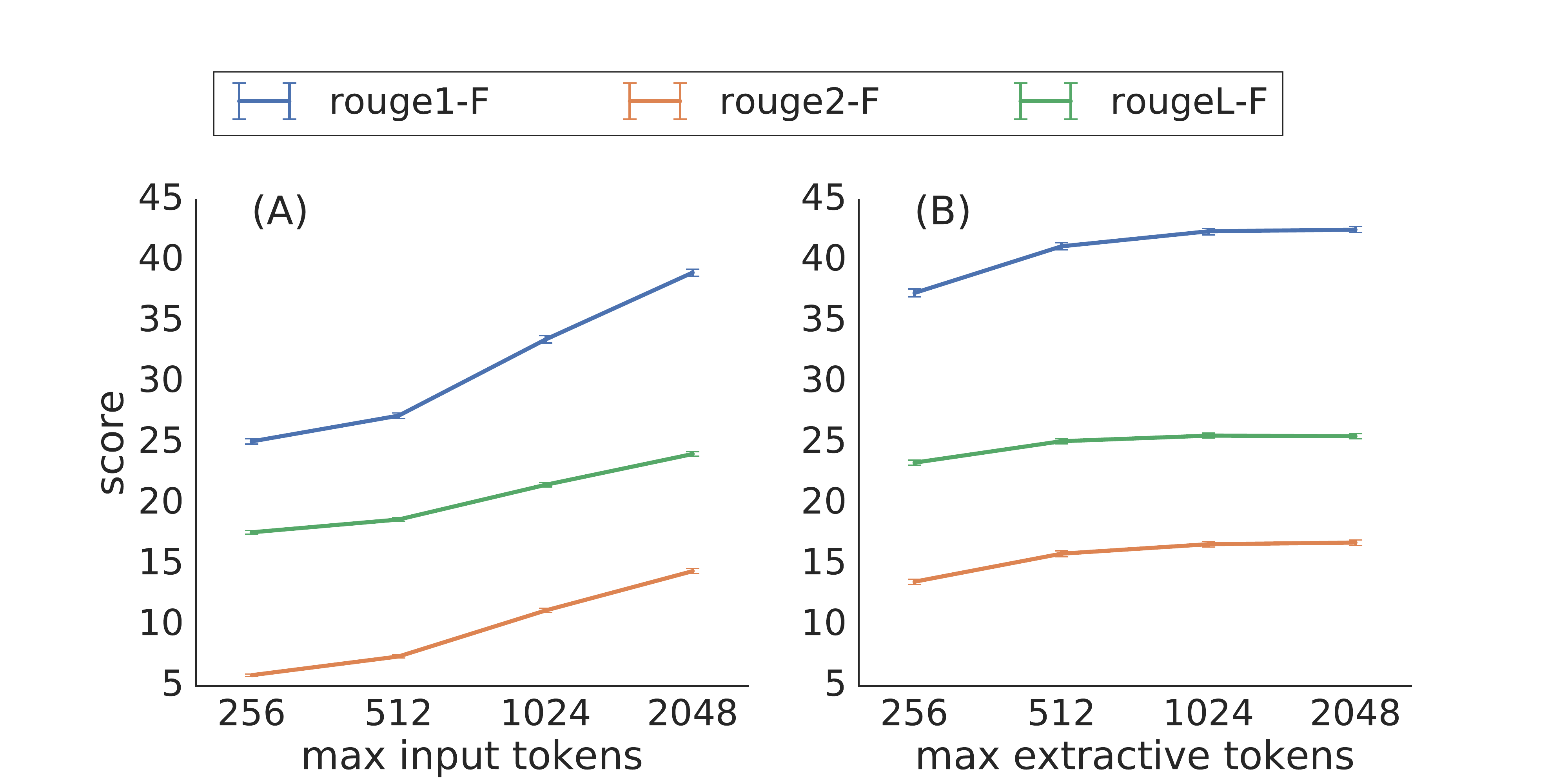}
\end{subfigure}
\hfill
\begin{subfigure}[t]{0.59\textwidth}
\includegraphics[trim=150 0 100 0,clip,width=\textwidth]{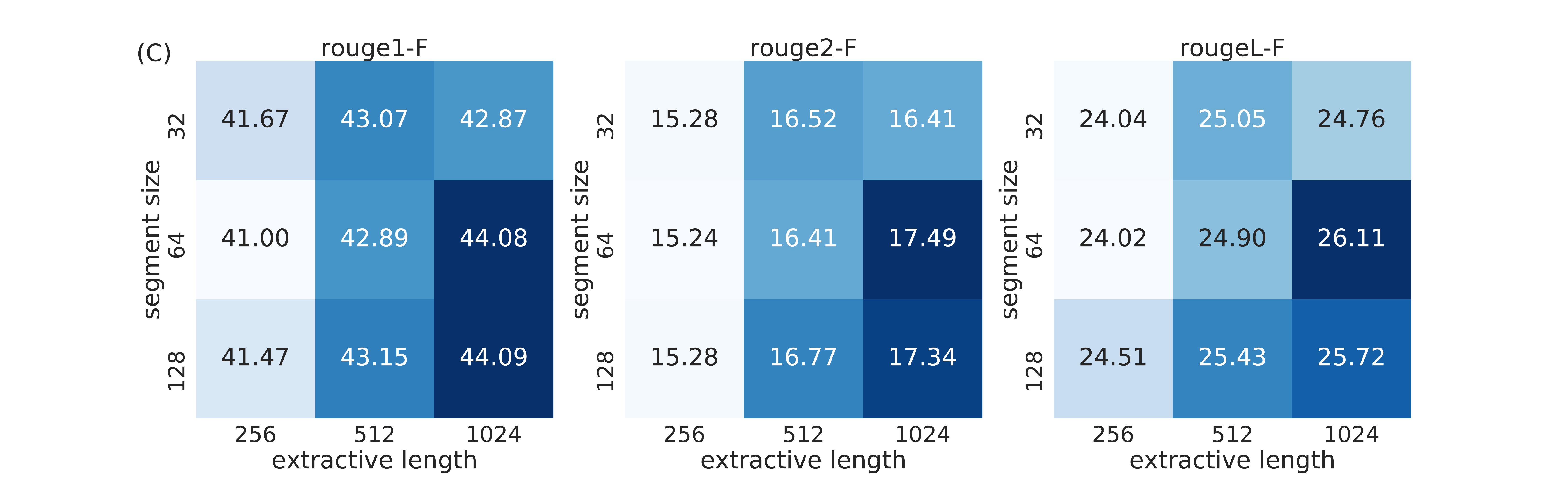}
\end{subfigure}
\caption{On the arXiv dataset, 
(a) Trunc models trained on different maximum input length, $L_{input}$.
(b) EA models trained on different maximum extractive length, $L_{ext}$. 
(c) Effect of segment length $L_{seg}$ and maximum extractive length $L_{ext}$  for SEAL model on the arXiv dataset.
\label{fig:input_len}
}
\end{figure}

\paragraph{Amount of input context} We show that the amount of input context the decoder attends to is important for LF-NAS. We limited the amount of input context to the Trunc model to $L_{input}$ to 256, 512, 1024 and 2048 tokens on the arXiv dataset. The performance of the model increases significantly as the length grows, Fig.~\ref{fig:input_len}(a), suggesting longer input context is crucial for LF-NAS. We observe similar trends on other LF-NAS datasets which is consistent with \citep{wikisum}. 

\paragraph{Input Snippets Selection} We show that the required amount of context the decoder attends to greatly reduces when better snippets are selected.
We trained EA models with  maximum extractive length $L_{ext}$ (Section~\ref{sec:ea}), of 256, 512, 1024 and 2048 tokens on the arXiv dataset.
With same number of tokens, the EA model achieves much better results compared to the Trunc model and plateaus at 1024 tokens, 
Fig.~\ref{fig:input_len}(B).

\paragraph{Methods of Creating Proxy Extractive Labels} We investigated methods to automatically create proxy labels based on multiple text similarity measures including ROUGE-1 or ROUGE-2 \citep{rouge}  precision/recall/F1 scores, and whether to create the labels sequentially  as in \citet{nallapati2016summarunner}. 
Appendix~\ref{appendix:rouge} shows that specific choice of labeling method doesn't make large differences. For simplicity, we chose non-sequential ROUGE-2-F in all of our models.

\paragraph{Segment Length and Extractive Length} For the SEAL model, 
when the segment length $L_{seg}$ is large, the decoder effectively attends to less context. 
On the other hand, when $L_{seg}$  is as small as a few words, the similarity labels become very spurious and are not meaningful to guide extraction.  Fig.~\ref{fig:input_len}(C) shows that larger maximum extractive lengths $L_{ext}$ are always better and 64-128 is the optimal segment length for the arXiv datasets. 
Therefore in all other experiments, we set the segment length to one-eighth of maximum decode length, $L_{seg} =L_{dec}/8$. Values of $L_{dec}$, can be found in Appendix~\ref{model_dims} on different dataset.

\begin{table*}[!htb]
\centering
\caption{Comparison of our models 
with state-of-the-art baselines. 
The metrics are, $R_1$: ROUGE-1 F1, $R_2$: ROUGE-2 F1, $R_L$: ROUGE-L F1. Best numbers for each dataset are bolded. There are two types of ROUGE-L scores, sentence-level and summary-level ~\citep{rouge}; * denotes the summary-level. $^{\ddagger}$ denotes pretrained model.}
\label{tab:models}
\small
\resizebox{0.8\columnwidth}{!}{

\begin{tabular}{cccccccccccccccccccc}
\hline
\multirow{2}{*}{Model} & \multirow{2}{*}{Trucated Input} & ArXiv & PubMed & Search2Wiki \\ 
 & & $R_1/R_2/R_L$ & $R_1/R_2/R_L$ & $R_1/R_2/R_L$ \\ \hline
DA\cite{arxiv} & Y & 35.8/11.1/31.8* & 38.9/15.4/35.2* & -\\ 
TLM\cite{tlm} & N & 42.4/15.2/24.1 & 41.4/15.9/24.3 & -\\ 
GLC\cite{xiao2019extractive} & N  & 43.6/17.4/29.1* & 44.9/19.7/31.4* & -\\
PEGASUS $^{\ddagger}$ \cite{zhang2019pegasus}  & Y  & \textbf{44.7}/17.3/25.8 & 45.5/19.9/27.7  & -\\
\hline
Trunc & Y & 32.9/11.3/20.7 & 39.8/15.5/24.3 & 29.1/14.5/24.2\\
CA & N & 42.3/16.0/25.4 & 43.5/16.4/26.3 & 34.6/19.3/28.5\\
EA & N & 43.0/17.3/25.6 & 45.8/19.9/\textbf{28.0} & 34.0/18.9/28.0\\
SEA & N & 44.3/\textbf{18.0}/\textbf{26.3}(\textbf{39.3*}) & \textbf{46.5}/\textbf{20.1}/\textbf{28.0}(\textbf{42.2*}) & \textbf{37.1}/\textbf{20.3}/\textbf{29.1}\\
\hline
\end{tabular}

}

\end{table*}

\paragraph{Comparison between Our Models and Other Works} We compare Trunc, CA, EA and SEAL models on arXiv, PubMed and Search2Wiki datasets in Table ~\ref{tab:models}.
The inputs to Trunc models are leading snippets on arXiv, PubMed datasets and top snippets ranked by tf-idf on Search2Wiki dataset,
similar to \citet{wikisum}. The inputs are concatenated and truncated to the first 1024 tokens.

Between our models, the CA and EA models have similar performance and consistently outperform the Trunc model on all datasets. The SEAL model performs better
than Trunc, CA and EA on long output datasets.
On Search2Wiki, which has larger number of training examples and longer input sequences, the advantage of the SEAL model is more obvious. 

Prior work on LF-NAS typically truncates the input or target in some way.
Truncating the targets\citep{arxiv, tlm} changes the nature of the problem, usually making it easier, and leading to higher evaluation metrics.
In this work, we ensure that the model's maximum input $L_{input}$ and decode $L_{dec}$ lengths (Appendix~\ref{model_dims}) exceed the 95\% percentile of the corresponding data distribution (Table~\ref{tab:data}), so that we closely tackle the intended summarization problem defined by the dataset.
The SEAL model has better performances comparing to previous state-of-the-art extractive \citep{xiao2019extractive} and abstractive \citep{tlm, arxiv} approaches on LF-NAS.

\section{Interpretability}
\label{sec:interp}
\definecolor{c1}{rgb}{0.6, 0.17, 0.31}
\definecolor{c2}{rgb}{0.26, 0.44, 0.56}
\definecolor{c3}{rgb}{0.45, 0.61, 0.0}
\definecolor{c4}{rgb}{0.6, 0.4, 0.2}
\begin{figure}[!htb]
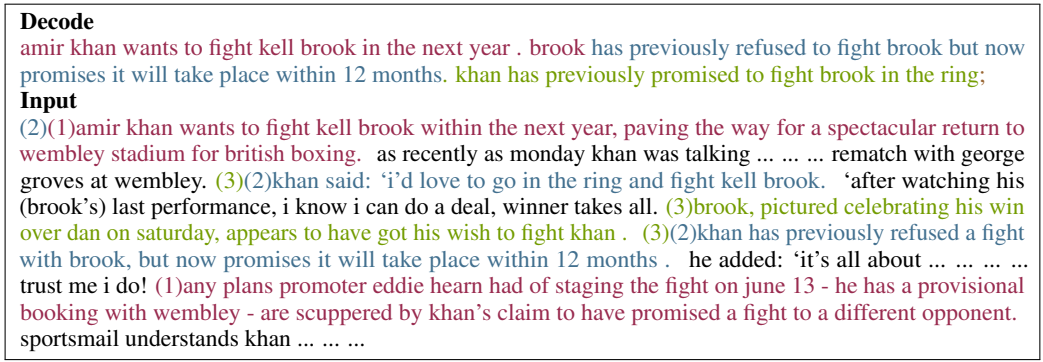

\fbox{
\centering
\begin{minipage}{38em}
\small
\textbf{Decode} \\
\textcolor{c1}{amir khan wants to fight kell brook in the next year . brook }\textcolor{c2}{has previously refused to fight brook but now promises it will take place within 12 months}\textcolor{c3}{ . khan has previously promised to fight brook in the ring}\textcolor{c4}{;}
\\ \textbf{Input} \\
\textcolor{c2}{(2)}\textcolor{c2}{\textcolor{c1}{(1)}\textcolor{c1}{amir khan wants to fight kell brook within the next year, paving the way for a spectacular return to wembley stadium for british boxing. 
}} as recently as monday khan was talking
 ... ... ...
 rematch with george groves at wembley. 
 \textcolor{c3}{(3)}\textcolor{c3}{\textcolor{c2}{(2)}\textcolor{c2}{khan said: ‘i’d love to go in the ring and fight kell brook. 
}} ‘after watching his (brook’s) last performance, i know i can do a deal, winner takes all. 
 \textcolor{c3}{(3)}\textcolor{c3}{brook, pictured celebrating his win over dan on saturday, appears to have got his wish to fight khan . 
} \textcolor{c3}{(3)}\textcolor{c3}{\textcolor{c2}{(2)}\textcolor{c2}{khan has previously refused a fight with brook, but now promises it will take place within 12 months . 
}} he added: ‘it’s all about
 ... ... ...
 ... trust me i do! 
 \textcolor{c1}{(1)}\textcolor{c1}{any plans promoter eddie hearn had of staging the fight on june 13 - he has a provisional booking with wembley - are scuppered by khan’s claim to have promised a fight to a different opponent. 
} sportsmail understands khan 
 ... ... ...
 \end{minipage}
}
\caption{Visualization of the SEAL model on an CNN/DailyMail example (best viewed in color). Segments of decodes are colored differently. Input snippets each segment attends to are are colored accordingly and the segment ids are inserted to the front. When multiple segment attend to the same input snippet, it is colored as the first segment.}
\label{fig:interp}
\end{figure}

The SEAL model provides a natural way to inspect what input sequence the decoder is likely using when generating a particular segment of the summary. In this section, we show how this enhances interpretability on an example from the CNN/DailyMail dataset (for ease of presentation), Fig~\ref{fig:interp}. In the Appendix we show more examples on long SDS and MDS datasets.

On CNN/DailyMail dataset, we find a SEAL model with ($L_{seg}=16$, $L_{snpt}=64$, $L_{ext}=256$, $N_{snpt}=16$) achieves R1/R2 of 39.3/16.5 which is on par with a Trunc model of 1024 input length.

When generating the first decode segment, the model attended to two snippets and copied from the summary-like lead sentence.
The second segment finishes the first sentence while continuously attending to the lead snippet, and begins a second sentence rephrasing three other snippets. The final segment writes a new sentence combining multiple snippets.

\section{Conclusion}

In this work, we studied and compared different LF-NAS models. We showed that models' performance heavily depends on seeing longer sequences from the input, and sparsely selecting better content relieves this requirement. 
We proposed the SEAL model, which encodes input snippets separately and dynamically selects sparse input snippets to attend to when generating different segments of the summary. The SEAL model achieves state-of-the-art results on existing LF-NAS datasets including arXiv, PubMed and outperform baseline models on our new, much longer dataset Search2Wiki.

\section*{Broader Impact}

This work may bring more attention to long document summarization research and spur new applications. If successful, producers of summaries from long/multiple documents may benefit from higher productivity  due to less manual work, while consumers may benefit from reduced information overload. 

A failure case of such a system is that it may generate text that is unfaithful to the source material, i.e. factually inaccurate, a risk that must be taken into account when deploying. The models biases present in the training data may also be reflected in the model output. 

\begin{ack}
We thank David Grangier for feedback and reviewing the manuscript and Ben Goodrich for helping with earlier iterations of the models.
\end{ack}

\bibliography{references}
\bibliographystyle{unsrtnat}

\newpage

\appendix

\section{Choice of Text Similarity Measure}
\label{appendix:rouge}
\begin{table}[!ht]
\centering
\begin{tabular}{cccccc}
\hline
\multicolumn{3}{c}{Self-supervision Labeling Method}  & \multicolumn{3}{c}{Metric} \\ \hline
\multicolumn{1}{c}{sequential} & \multicolumn{1}{c}{ngram} & \multicolumn{1}{c}{type} & \multicolumn{1}{c}{R1} & \multicolumn{1}{c}{R2} & \multicolumn{1}{c}{RL} \\ \hline
\multirow{6}{*}{False} & \multirow{3}{*}{1} & f1 & 41.30 & 16.27 & 25.26 \\
 \cline{3-6} 
 &  & precision & 41.85 & 16.50 & 25.38 \\
 \cline{3-6} 
 &  & recall & 40.93 & 15.92 & 24.87 \\
 \cline{2-2}  \cline{3-6} 
 & \multirow{3}{*}{2} & f1 & 41.29 & 16.46 & 24.89 \\
 \cline{3-6} 
 &  & precision & 41.75 & 16.67 & 25.48 \\
 \cline{3-6} 
 &  & recall & 41.27 & 16.40 & 24.95 \\
 \cline{1-1}  \cline{2-2}  \cline{3-6} 
\multirow{6}{*}{True} & \multirow{3}{*}{1} & f1 & 41.97 & 16.64 & 25.31 \\
 \cline{3-6} 
 &  & precision & 40.30 & 15.84 & 24.28 \\
 \cline{3-6} 
 &  & recall & 41.84 & 16.67 & 25.32 \\
 \cline{2-2}  \cline{3-6} 
 & \multirow{3}{*}{2} & f1 & 40.80 & 16.24 & 24.47 \\
 \cline{3-6} 
 &  & precision & 41.04 & 16.22 & 24.70 \\
 \cline{3-6} 
 &  & recall & 40.82 & 16.26 & 24.47 \\
 \hline
\end{tabular}
\caption{Comparison of EA models trained with different self-supervised labeling method on arXiv dataset using the extractive abstractive model. }
\label{tab:rouge}
\end{table}

\section{Search2Wiki Clone Detection}
\label{web2wiki_clone}
In WikiSum \citep{wikisum}, to detect whether a source document, $d$, is a clone of a Wikipedia article, $a$, the maximum recall of unigrams between each section of $a$ and $d$ is computed as follows:
\[
r(d, a) = \max_{s \in sections(a)} \frac{|unigrams(d) \cap unigrams(s)|}{|unigrams(s)|} \]

A clone is detected if $r(d, a)>0.5$. While this approach detects and filters most of the clones, we observed many near-clone documents left undetected in the WikiSum dataset. Most of these near-clones are documents that copy small parts of the Wikipedia article rather than the whole article or a whole section. To filter these near-clones more effectively and make the dataset more abstractive, we extended the equation above to a maximum recall of $n$-grams between each section of $a$ and $d$ as follows:
\[
r(d, a, n) = \max_{s \in sections(a)} \frac{|n-grams(d) \cap n-grams(s)|}{|n-grams(s)|} \]

We experimented with different values of $n$. A near-clone is detected in Search2Wiki if $r(d, a, 6)>0.2$.

\section{Model Dimensions}
\label{model_dims}
\begin{table*}[!h]
\centering 
\begin{tabular}{c|ccc|c|ccc} 
dataset       & $L_{input}$ & $L_{snpt}$ & $N_{snpt}$ & $L_{ext}$ & $L_{dec}$ & $L_{seg}$ & $N_{seg}$ \\ \hline
arXiv         & 28k       & 128         & 224             & 1024 & 768        & 96          & 8            \\
PubMed        & 12k       & 128         & 96              & 1024 & 512        & 64          & 8            \\
Search2Wiki & 100k      & 128         & 768             & 1024 & 512        & 64          & 8            \\
\hline
\end{tabular}
\caption{Dimensions of the models inputs and outputs for CA, EA and SEAL. Length are all in unit of subword tokens.
$L_{input}$ is the maximum input length. $L_{snpt}$ is the snippet length. $N_{snpt}$ is the maximum number of snippets. $L_{input} = L_{snpt} \times N_{snpt}$. $L_{ext}$ is the maximum extractive length. $L_{dec}$ is the maximum decode length. For the SEAL models, $L_{seg}$ is the decode segment length and $N_{seg}$ is the maximum number of segments.$L_{dec} = L_{seg} \times N_{seg}$.
}
\label{tab:dim}
\end{table*}






\end{document}